\documentclass[letterpaper,10pt,conference]{ieeeconf}  

\IEEEoverridecommandlockouts                              

\usepackage{booktabs}
\usepackage{url}
\usepackage{amsmath,amssymb}
\usepackage{amsfonts}
\usepackage{bbold}
\usepackage{xcolor}
\usepackage[english]{babel}
\usepackage{graphicx}
\usepackage{subcaption}
\usepackage{algorithm}
\usepackage{comment}
\usepackage{float}
\usepackage{hyperref}
\usepackage[noend]{algpseudocode}

\newcommand{\rev}[1]{{#1}}

\renewcommand{\baselinestretch}{0.97}

\title{\LARGE \bf
Cooperative Object Transportation using Gibbs Random Fields
}

\author{Paulo Rezeck$^{1}$ and Renato M. Assunção$^{1,2}$ and Luiz Chaimowicz$^{1}$
\thanks{Paulo Rezeck, Renato M. Assunção and Luiz Chaimowicz are with the Department of Computer Science, Universidade Federal de Minas Gerais,
        Brazil$^{1}$.
        Renato M. Assunção is also with the ESRI Inc., Redlands, CA$^{2}$.
        Email: {\tt\small \{rezeck,assuncao,chaimo\}@dcc.ufmg.br}.
        This work was partially supported by CAPES, CNPq, and Fapemig.}%
}

\begin{document}

\maketitle
\thispagestyle{empty}
\pagestyle{empty}


\begin{abstract}  

This paper presents a novel methodology that allows a swarm of robots to perform a cooperative transportation task. Our approach consists of modeling the swarm as a {\em Gibbs Random Field} (GRF), taking advantage of this framework's locality properties. By setting appropriate potential functions, robots can dynamically navigate, form groups, and perform cooperative transportation in a completely decentralized fashion. Moreover, these behaviors emerge from the local interactions without the need for explicit communication or coordination. To evaluate our methodology, we perform a series of simulations and proof-of-concept experiments in different scenarios. Our results show that the method is scalable, adaptable, and robust to failures and changes in the environment.

\end{abstract}

\section{Introduction}
\label{sec:intro}

Robotic swarms are composed of a large number of robots that generally rely on emergent collective behaviors to solve complex problems. Such systems present desirable characteristics, such as robustness, adaptability, simplicity, and scalability, which are important in different tasks~\cite{tarapore2020sparse},~\cite{schranz2020swarm}.

In particular, robotic swarms capable of cooperatively transporting objects may be suitable for many applications with high societal and economic impact potential. For instance, one may use robotic swarms for operations where the use of sophisticated robots is impossible or impractical, such as warehouse automation, waste disposal, and demining.

Despite several advantages of using robotic swarms robots for cooperative transportation, designing decentralized control methods for such application is not simple. Some of the main challenges consist of aligning and synchronizing the forces applied to sustain the transportation. The method must be robust to objects of different shapes and resilient to changes in the environment, such as surfaces with different coefficients of friction, and robot failures.

\begin{figure}[t]
    \centering
    \includegraphics[width=0.40\textwidth]{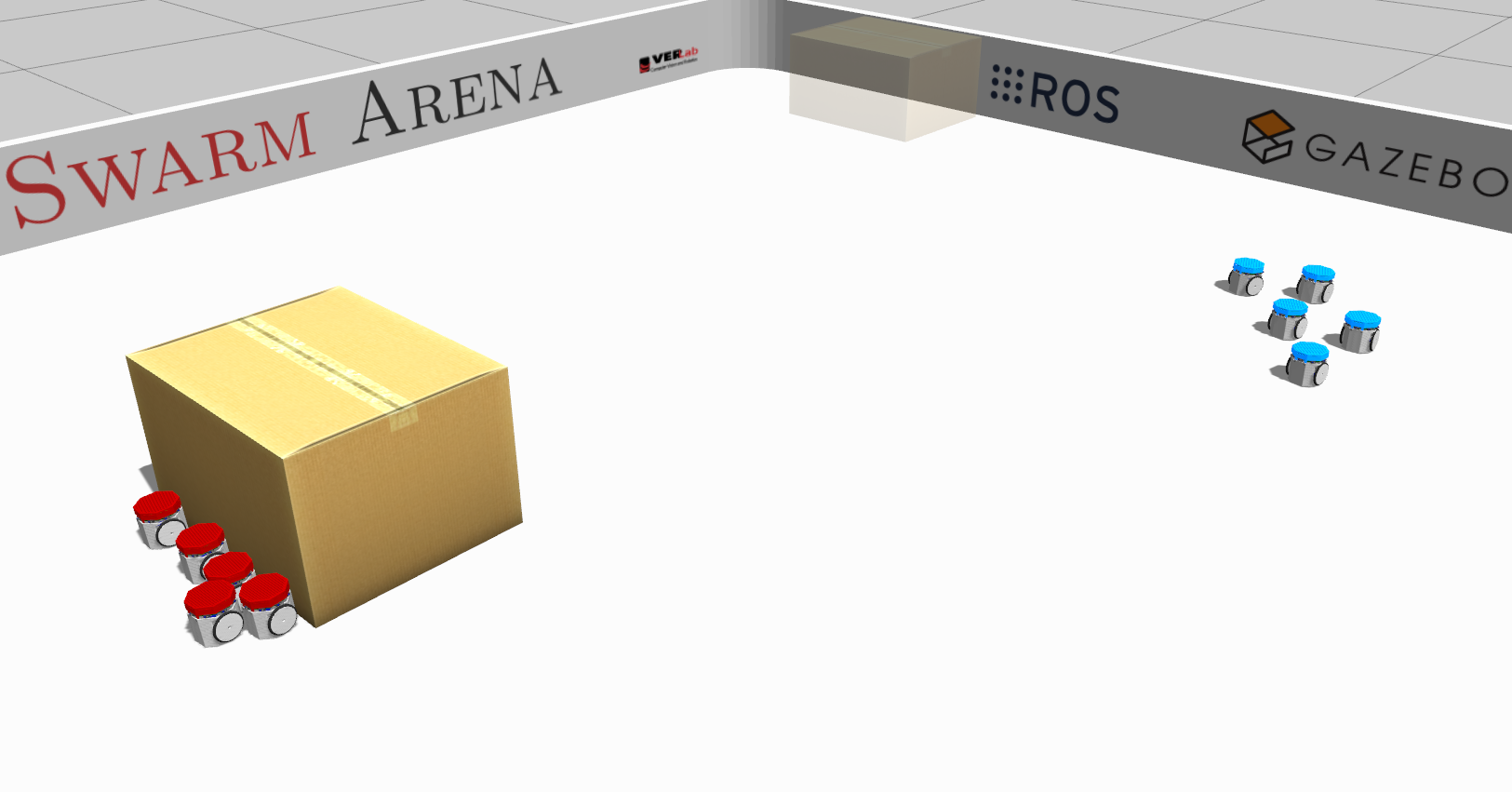}
    \caption{Swarm of robots cooperatively transporting an object (solid cardboard) toward a goal (transparent cardboard). 
    }
    \label{fig:transport_example}
\end{figure}

This work presents a novel stochastic and decentralized approach that allows a swarm of robots to navigate autonomously through a bounded environment and cooperatively transport an object toward its goal location, as shown in Fig.~\ref{fig:transport_example}. Our approach consists of modeling the robotic swarm as a Gibbs Random Field (GRF) and defining its potential energy as a combination of the Coulomb-Buckingham potential and kinetic energy. The Coulomb-Buckingham potential enables the robots to aggregate, interact with the object by pushing it, and avoid obstacles in the environment. The kinetic energy allows the robots to reach a consensus on their relative velocities concerning their neighborhood and circulate the object looking for adequate pushing positions.  

Dynamically and autonomously adapting the potential and energy parameters, robots can navigate through the environment looking for the object to be transported, form groups, and push the object toward a goal location. These behaviors emerge from the local interactions without the need for explicit communication or coordination. The robots only need to be able to estimate the relative position and velocity of their neighbors \rev{and also distinguish between obstacles and the object detected within their sensing range.} Moreover, the robots do not need any information about the object (i.e., location, size, mass, and shape), except for its goal location.

%
\begin{table*}[!h]
\caption{Main characteristics of research works using pushing-only transport strategy.}
\label{tab:comparision}
\resizebox{\textwidth}{!}{\begin{tabular}{llccccccccccccc}
\toprule
\multicolumn{1}{c}{\textbf{Studies}} & \multicolumn{1}{c}{\textbf{Control method}} & & \multicolumn{4}{c}{\textbf{Experimental setup}} & & \multicolumn{1}{c}{\textbf{Scalability}} & \multicolumn{2}{c}{\textbf{Adaptability test}} & & \multicolumn{3}{c}{\textbf{Robustness test}} \\ \cmidrule{4-7} \cmidrule{10-11} \cmidrule{13-15} 
\multicolumn{1}{c}{\textbf{\begin{tabular}[c]{@{}c@{}}\end{tabular}}} & &
\multicolumn{1}{c}{} & \multicolumn{1}{c}{\textbf{\begin{tabular}[c]{@{}c@{}}Number of \\ robots\end{tabular}}} & 
\multicolumn{1}{c}{\textbf{\begin{tabular}[c]{@{}c@{}}Env. with \\ obstacles\end{tabular}}} & \multicolumn{1}{c}{\textbf{\begin{tabular}[c]{@{}c@{}}Transport \\ direction\end{tabular}}} & \multicolumn{1}{c}{\textbf{\begin{tabular}[c]{@{}c@{}}Random \\ position\end{tabular}}} & &
\multicolumn{1}{c}{\textbf{\begin{tabular}[c]{@{}c@{}}\end{tabular}}} & 
\multicolumn{1}{c}{\textbf{\begin{tabular}[c]{@{}c@{}}Robot\\ failure \end{tabular}}} & 
\multicolumn{1}{c}{\textbf{\begin{tabular}[c]{@{}c@{}}Change in the\\ environment\end{tabular}}} & &
\multicolumn{1}{c}{\textbf{\begin{tabular}[c]{@{}c@{}}Object\\ mass\end{tabular}}} & 
\multicolumn{1}{c}{\textbf{\begin{tabular}[c]{@{}c@{}}Object\\ shape\end{tabular}}} &
\multicolumn{1}{c}{\textbf{\begin{tabular}[c]{@{}c@{}}Object\\ size\end{tabular}}} \\ \midrule
Kube and Bonabeau, 2000~\cite{kube2000cooperative} & Behavior-based  & & 2-20  & no & towards a goal & no & & high & no & no & & no & no & yes\\
Yamada and Saito, 2001~\cite{yamada2001adaptive} & Behavior-based  & & 4  & no & towards a goal & no & & low & yes & yes & & no & no & no\\
Fujisawa et al., 2013~\cite{fujisawa2013cooperative} & Behavior-based  & & 4-10  & no & towards a goal & yes & & high & yes & yes & & yes & no & no\\
Jianing Chen et al., 2015~\cite{chen2015occlusion} & Behavior-based  & & 20  & yes & towards a goal & yes & & high & no & no & & yes & yes & yes\\
Alkilabi et al., 2018~\cite{alkilabi2018evolving} & Neural Networks  & & 2-6  & no & towards a goal & no & & high & no & no & & yes & no & yes\\
\textbf{Ours} & \textbf{GRFs} & & \textbf{2-20}  & \textbf{yes} & \textbf{towards a goal} & \textbf{yes} & & \textbf{high} & \textbf{yes} & \textbf{yes} & & \textbf{yes} & \textbf{yes} & \textbf{yes} \\ 
\bottomrule
\end{tabular}}
\end{table*}

\section{Related Work}
\label{sec:relatedwork}
Over the years, a wide range of control and coordination methods have been proposed to perform cooperative object transportation using multiple robots.
Recently, Tuci et al.~\cite{tuci2018cooperative} presented a review of the state-of-the-art works in this area, which categorizes the most common transportation strategies into three types: pushing, caging, and grasping. 

Pushing strategies, such as those used in this work, consider that the robots are not attached to the object. They provide a simple way of manipulating relatively large objects and do not require any sophisticated mechanism to apply forces to the object. Pushing is also interesting because, unlike other strategies, it allows robots to aggregate and apply forces to specific points on the object, enabling the transportation of large and heavy objects when the number of robots is increased. Moreover, it allows the use of simpler robots, providing a suitable scenario for robotic swarms.

While there is a large body of work in cooperative object transport~\cite{tuci2018cooperative}, we focus our discussion on approaches that do not require centralized planners or direct communication among robots to push an object from a random location to a specific goal. Table~\ref{tab:comparision} summarizes the main characteristics of these approaches, and in the following, we review and compare these works with ours, mainly discussing scalability, adaptability, and robustness issues. By adaptability, we consider the team's ability to work reliably even with individual robot failures or changes in the environment, while robustness is related to the power of the methodology to deal with objects of different mass, size, or shape.  

Kube et al.~\cite{kube1993collective} presented one of the first studies that formally dealt with the dynamics of cooperative transport. The authors demonstrated that coordinated efforts produced by a homogeneous group of simple robots pushing an object are possible without using a direct communication mechanism. 
Further, in Kube and Zhang~\cite{kube1997task}, the authors extended the original work by constructing a new robot for experimentation. Also, In Kube and Bonabeau~\cite{kube2000cooperative}, they proposed the addition of a stagnation recovery strategy avoiding deadlock conditions in which the robots cancel the pushing forces on the object depending on where they are positioned around it. Moreover, they extended their approach, allowing the robots to push the object towards a fixed goal position and performed experiments to test their method's robustness and scalability.

Yamada and Saito~\cite{yamada2001adaptive} also presented theoretical and experimental analyses to support the assumption that robots can transport objects without using direct communication. Unlike previous works, the authors demonstrated that the use of indirect communication also allows the transport of an object towards a goal location considering a dynamic environment. 
Their approach consists of using an action selection method to design a behavior-based control method that is robust to a small change in the environment, such as increase the number of robots or the mass of the object. 
To evaluate the performance of their approach, the authors executed real experiments using four robots showing that the robots can operate in a simple environment where individual robots are required to push a light object or in complex environments where multiple robots are required to push a heavy object cooperatively. Although their approach allows adaptability to small changes in the environment, the authors did not evaluate the robustness of the method to objects with different shapes, mass, and sizes. Furthermore, the method does not seem to scale easily as the system grows. 

Fujisawa et al., 2013~\cite{fujisawa2013cooperative} presented a cooperative transport approach that uses an interesting mechanism for indirect communication via artificial pheromones as seen in ants. Such a mechanism allows the robots to sense and lay on the terrain a volatile alcohol substance, mimicking pheromone during trail formation. The authors also proposed a behavior-based algorithm using a deterministic finite automaton, which allows the robots to perform a random search to find a food item (i.e., a heavy object) and transport it to a goal location (i.e., the nest). To evaluate the efficiency of the proposed system, they perform experiments with up to 10 robots demonstrating the adaptability of their approach but did not evaluate the system's scalability and the robustness of the method as to the object's shape and size changes.  

Jianing Chen et al.~\cite{chen2015occlusion} proposed another behavior-based strategy for cooperative transport that deals with goal occlusion by keeping the robots pushing the object even when not detecting the goal.
They performed experiments using twenty robots to transport objects of different shapes toward their goals and provided analytical proof of the method's effectiveness. Also, the authors demonstrated other interesting experiments in which they consider the goal location as a mobile robot remotely controlled by a human, showing potential applications in the context of human-robot interaction. The work was further extended with this focus in Kapellmann-Zafra et al.~\cite{kapellmann2016using}. Although the authors demonstrated the system's scalability and robustness to different types of objects, tests evaluating the adaptability of the system to failures or changes in the environment were not performed.

Differently from previous approaches, Alkilabi et al.~\cite{alkilabi2017cooperative} proposed an approach based on recurrent neural networks and evolutionary algorithms, which allow a group of robots with a minimalist sensory apparatus to transport a heavy object by perceiving its movement. The best instance of the evolved controller was extensively tested on physical robots to transport objects of different sizes and masses in an arbitrary direction. Later on, Alkilabi et al.~\cite{alkilabi2018evolving} presented a complementary study extending their neural-controller with mechanisms to direct the transport towards a specific target location. The authors showed that the transport strategies are scalable concerning the group size and robust enough to deal with boxes of various masses and sizes. Despite presenting interesting results in a large set of experiments, the strategy still requires robots to start with a direct view of the object. Also, it does not consider objects with different shapes or obstacles in the environment.

This paper presents an approach that allows a swarm of robots to coordinate their motion to navigate a bounded environment and cooperatively transport an object toward its goal location. We model the swarm as a Gibbs Random Fields (GRF) and take advantage of this framework's properties to control the robots in a completely decentralized fashion, relying only on local information. Different from other methods, we do not require pre-synthesized behaviors or automatic learning methods. Using appropriate potential functions, the desired behaviors emerge directly from local interactions, bringing scalability, robustness, and adaptability to our method.

\section{Methodology}
\label{sec:methodology}

The proposed approach is based on our recent work~\cite{rezeck2021flocking} in which we proposed and evaluated a novel methodology that allows a swarm of heterogeneous robots to simultaneously achieve flocking and segregative behaviors.
Here, we adapt and extend this methodology allowing a swarm of robots to cohesively navigate through a bounded environment and cooperatively transport an object toward its goal location. 

The main idea behind our approach consists of modeling the swarm as a dynamic \textit{Gibbs Random Field} (GRF) and then sampling velocities for each robot in a decentralized way, which leads the entire swarm to converge toward the global minimum of the potential. Thus, setting the potential functions accordingly, we can make the swarm execute the desired task in a completely decentralized fashion.


\subsection{Formalization}
\label{subsec:formalization}
Assume a set $\mathcal{R}$ of $\eta$ homogeneous robots navigating through a bounded region within the two-dimensional Euclidean space.
The state of the $i$-th robot at time step $t$ is represented by its pose $\mathbf{q}_i ^{(t)}$ and velocity $\displaystyle{\dot{\mathbf{q}_i}^{(t)} = \mathbf{v}_i^{(t)}}$, which is bounded by $\displaystyle{||\mathbf{v}_i^{(t)}|| \leq v_{max}}$. The robots are driven by a kinematic model with motion model defined as $\displaystyle{\mathcal{K} : (\mathbf{q}_i^{(t)}, \mathbf{v}_i^{(t)}) \rightarrow (\mathbf{q}_i^{(t+1)})}$ and have a circular sensing range of radius $\lambda$, enabling them to estimate the relative position and velocity of other robots as well as objects and obstacles within this range. 

We represent an object as a finite set of points $\displaystyle{\mathcal{O} = \{\mathbf{o}_1, ..., \mathbf{o}_n\}}$ outlining the object perimeter. An object detected by the $i$-th robot within its circular sensing range consists of a subset of points $\mathcal{O}_i \subset \mathcal{O}$, where $\displaystyle{\mathbf{o}_j \in \mathcal{O}_i \rightarrow ||\mathbf{o}_j - \mathbf{q}_i|| \leq \lambda}$ and $\displaystyle{||\mathbf{o}_j - \mathbf{q}_i||}$ is the Euclidean norm between two points. The robot represents obstacles in the same way as the object. 
Thus, we require the $i$-th robot to be able to arrange the points detected on objects and obstacles at time $t$ into two distinct sets, $\mathcal{O}^{(t)}_i$ and $\mathcal{W}^{(t)}_i$, respectively. In addition, we assume that the robots can track the goal location, $\mathbf{g}$, of object $\mathcal{O}$ from any point within the environment. 


\subsection{Applying GRFs concepts to swarm robotics}
\label{subsec:grf_extension}

A GRF is a probabilistic graphical model that is a particular case of a  \textit{Markov Random Field} (MRF) when the joint probability density of the random variables is represented by the Gibbs Measure~\cite{kindermann1980markov}. Such models are conditioned by the Markov properties~\cite{koller2009probabilistic},
which are convenient to model robotic swarms. Such properties imply the conditional independence of information coming from outside a neighborhood system, supporting the requirement of local interactions. 


\subsubsection{Modeling a swarm of robots}
assume an undirected graph $\displaystyle{\mathbf{G} = (\mathcal{R},\mathbf{E})}$ with each vertex representing one of the robots. A random field on $\mathbf{G}$ consists of a collection of random variables $\mathbf{X} = \{X_i\}_{i \in \mathcal{R}}$ and, for each $i \in \mathcal{R}$, let $\Lambda_i$ be a finite set called the phase space for the $i$-th robot. A configuration of the system $\mathbf{X}$ at time step $t$ is defined as $\displaystyle{\mathbf{x}^{(t)} = \{\mathbf{v}_1,... ,\mathbf{v}_\eta\}}$, where $\mathbf{v}_i \in \Lambda_i$ and represents the velocities performed by each robot at that time step.

A neighborhood system on $\mathcal{R}$, given a configuration $\mathbf{x}^{(t)}$, is a family $\mathcal{N} = \{\mathcal{N}_i\}_{i \in \mathcal{R}}$, where $\mathcal{N}_i \subset \mathcal{R}$ is the set of neighbors for the $i$-th robot. The neighborhood is constrained by the sensing range $\lambda$ satisfying $\displaystyle{\mathcal{N}_i \triangleq \{j \in \mathcal{R}: j \neq i, ||\mathbf{q}_j - \mathbf{q}_i|| \leq \lambda \}}$, and 
 induces the configuration of the graph $\mathbf{G}$ by setting an edge between two robots if they are neighbors. 


A random field $\mathbf{X}$ is called a GRF if the joint probability density of the system is represented by,
\begin{equation}
    P(\mathbf{X} = \mathbf{x}) = \frac{1}{Z}e^{-\frac{H(x)}{T}}, \textrm{ with } Z = \sum \limits_{z} e^{-\frac{H(z)}{T}},
    \label{eq:gibbsdistribution}
\end{equation}
where $Z$ is a normalizing term; $T$ is a constant interpreted as the temperature in the statistical physics context and taken as equal to $1$ in this paper; $\frac{1}{Z}e^{-\frac{H(x)}{T}}$ is called the Gibbs distribution; and $H(x)$ is the potential energy of the system.

In short, the distribution function~(\ref{eq:gibbsdistribution}) establishes that the probability of the random field $\mathbf{X}$ assuming state $\mathbf{x}$ is proportional to the potential energy of such a state divided by the sum of the potential energy relative to all states $z$ that the field can assume. It implies that the low energy states are more likely than those of higher energy.

The potential energy $H(x)$ consists of the summation of values produced by potential functions, $U_A : \Lambda \rightarrow \mathbb{R}$, which quantifies the state $\mathbf{x}$ of the swarm system concerning a physical property or behavior. Formally, it is defined by pairwise interactions between neighboring vertices as
\begin{equation}
H(x) = \sum \limits_{i \in \mathcal{R}} U_{\{i\}}(\mathbf{v}_i) + \sum \limits_{(i,j) \in \mathcal{R}  \times \mathcal{R}, j \in \mathcal{N}_i} U_{\{i,j\}}(\mathbf{v}_i, \mathbf{v}_j),
\label{eq:potentialenergy2}
\end{equation}
where $U_{\{i\}}(\mathbf{v}_i)$ is interpreted as the potential for the $i$-th robot to reach the velocity $\mathbf{v}_i$, and $U_{\{i,j\}}(\mathbf{v}_i, \mathbf{v}_j)$ is the potential regarding the velocities of the neighboring $(\mathbf{v}_i, \mathbf{v}_j)$ pair of vertices.

\subsubsection{Parallel Gibbs sampling}
the probability function~(\ref{eq:gibbsdistribution}) enables one to compute the probability of the entire swarm reaching a certain configuration $\mathbf{x}$. However, what we require is the sampling of each \rev{robots\textquotesingle~velocity} in a decentralized way, given only the neighborhood of the robots. 

Parallel Gibbs sampling is a strategy that simultaneously updates the \rev{robots\textquotesingle~velocities} based on their configuration at time $t$. Such strategy is possible due to the local nature of the potential energy~($\ref{eq:potentialenergy2}$), and it implies that we do not require the knowledge of the entire swarm to sample velocities for the $i$-th robot, but only information about its $\mathcal{N}_i$ neighbors. Indeed, by replacing (\ref{eq:potentialenergy2}) in (\ref{eq:gibbsdistribution}), we obtain
\begin{equation}
    P_i(\mathbf{v_i}, \mathbf{\bar v_i}|\mathbf{x}) = 
    \frac{e^{-\left(U_{\{i\}}(\mathbf{\bar v_i}) + \sum \limits_{\forall j \in \mathcal{N}_i}U_{\{i,j\}}(\mathbf{\bar v_i}, \mathbf{v_j})\right)}}
    {\sum \limits_{\mathbf{z_i} \in \mathbf{Z}_i(x)} e^{-\left(U_{\{i\}}(\mathbf{z_i}) + \sum \limits_{\forall j \in \mathcal{N}_i}U_{\{i,j\}}(\mathbf{z_i}, \mathbf{v_j})\right)}},
    \label{eq:problocal}
\end{equation}
where $\mathbf{Z}_i(\mathbf{x}) \triangleq \{\mathbf{z}_i : ||\mathbf{z}_i|| \leq v_{max}\}$ is a set of possible velocities for the $i$-th robot given the configuration $\mathbf{x}^{(t)}$, restricted by  $\mathbf{Z}_i(\mathbf{x}) \subset \Lambda_i$; and $\mathbf{\bar v_i}$ is a velocity sampled in $\mathbf{Z}_i(\mathbf{x})$ representing a likely velocity for the next time step, $\displaystyle{\mathbf{v_i}^{(t+1)} = \mathbf{\bar v_i}}$.

We use (\ref{eq:problocal}) to sample appropriate velocities for each robot using only the current state and information about their neighborhood. A common approach to sampling velocities given the probability function~ (\ref{eq:problocal}) consists of using Markov Chain Monte Carlo (MCMC) methods. In this work, we use the Metropolis-Hastings algorithm~\cite{peskun1973optimum}.

\subsubsection{Swarm behaviors and potential functions}
one of the many advantages of modeling a robotic swarm as a GRF is the flexibility to implement different swarm behaviors, changing only the potential functions used by the potential energy~(\ref{eq:potentialenergy2}). We extend the definition of the potential $U_{\{i\}}(\mathbf{v}_i)$ to allow the robots to interact with objects as well as to avoid collisions with obstacles within the environment. Also, the potential $U_{\{i,j\}}(\mathbf{v}_i, \mathbf{v}_j)$ is used to maintain the robots cohesively navigating through the environment. 
For this, we use the Coulomb-Buckingham potential in conjunction with a kinetic energy term. 

\begin{figure}[t]
		\centering
		\includegraphics[width=.575\columnwidth]{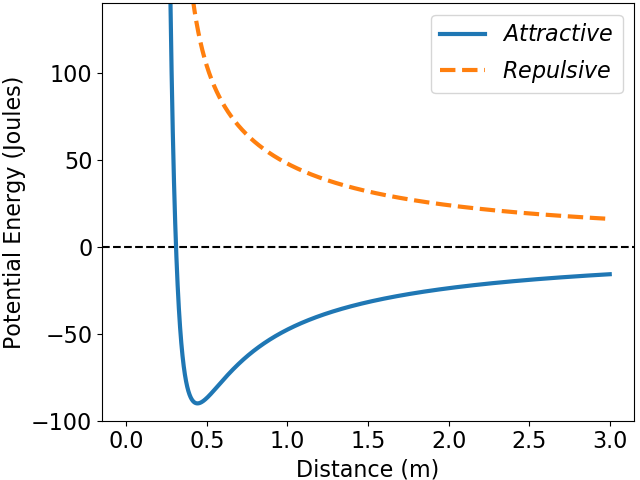}
	\caption{The Coulomb-Buckingham potential function depends on the distance $r$ among two particles $i$ and $j$ and their charge $c_i$ and $c_j$. We set the charge to produce attractive behavior among the robots and with the object, or repulsive behaviors to avoid collision with obstacles~\cite{rezeck2021flocking}.
	} 
	\label{fig:cbpotential}
\end{figure}

The Coulomb-Buckingham potential~\cite{buckingham1938classical} is a combination of the Lennard-Jones potential with the Coulomb potential. It describes the interaction among particles considering the distance between them and their charges. We take advantage of such a potential to model the cohesion among the robots as well as their interaction with objects and obstacles. Fig.~\ref{fig:cbpotential} illustrates the Coulomb-Buckingham potential, given by:
\begin{equation}
\Phi (r)=\varepsilon \left({\frac {6}{\alpha -6}}e^{\alpha} \left(1-{\frac {r}{r_{0}}}\right)-{\frac {\alpha }{\alpha -6}}\left({\frac {r_{0}}{r}}\right)^{6}\right) + \frac {c_{i}c_{j}}{4\pi \varepsilon_0 r},
    \label{eq:cbpotential}
\end{equation}
where $r = ||\mathbf{q}_j - \mathbf{q}_i||$ is the euclidean distance between the particles $i$ and $j$; $\varepsilon$ is a constant representing the depth of the minimum energy; $r_{0}$ is the minimum energy distance; $\alpha$ is a free dimensionless parameter; $c_i$ and $c_j$ are the charges of the particles $i$ and $j$; and $\varepsilon_0$ is an electric constant. 

We assume classical mechanics to compute the kinetic energy regarding the robots and object motion. Kinetic energy allows the robots to maintain a consensus on their velocities while navigating as a group and also to move around the object. Formally, assume $\mathbf{V}$ as the sum of the velocity vectors and $m$ is the mass of the system. The kinetic energy $\mathbf{E}$ is defined as
\begin{equation}
 \mathbf{E}(\mathbf{V}) = \frac{1}{2} m (\mathbf{V} \cdot \mathbf{V}).
\end{equation}

\begin{figure*}[t]
		\centering
		\includegraphics[width=1.8\columnwidth]{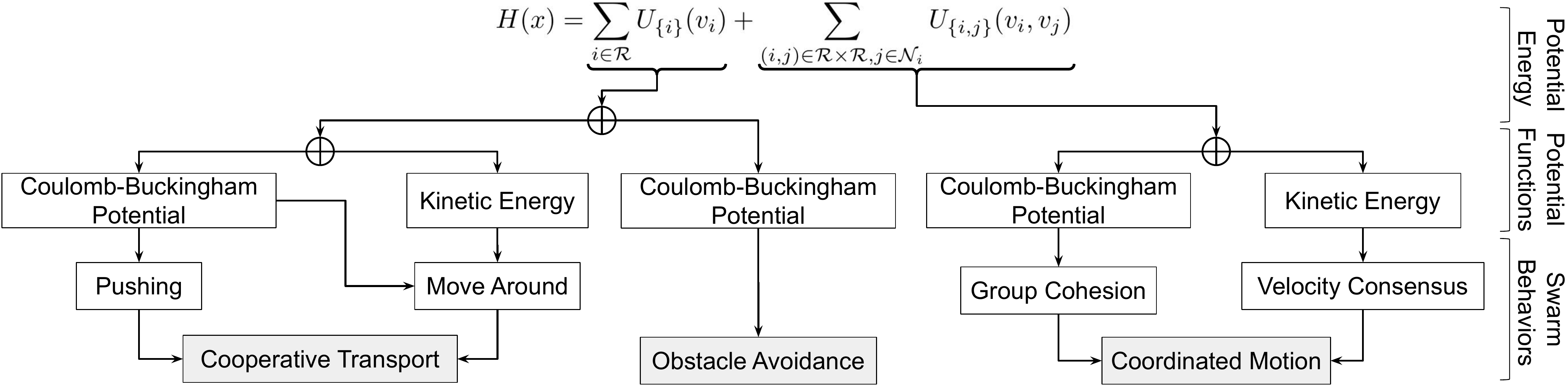}
	\caption{Diagram showing how the swarm behaviors emerge from the combination of the potential functions.} 
	\label{fig:diagram}
\end{figure*}

Fig.~\ref{fig:diagram} presents a diagram illustrating how the desired behaviors emerge from the combination of Coulomb-Buckingham potential and kinetic energy. These behaviors are detailed in the next sections.

\subsection{Cooperative transport}
\label{subsec:cooperative_transport}

For the cooperative transportation task, we design the potential $U_{\{i\}}(\mathbf{v}_i)$ to simultaneously generate the move around and pushing behaviors. The pushing behavior comes directly from the Coulomb-Buckingham potential, and the move around comes from the combination of such potential with the kinetic energy. We note that increasing or decreasing the Coulomb-Buckingham potential implies favoring the pushing or moving around behavior. Thus, we propose a conditional factor: if the robot is in a position where it can push the object towards its goal, the robot increases the Coulomb-Buckingham potential. Otherwise, it decreases.

The {\em Move Around Behavior} comes from the combination of the Coulomb-Buckingham potential with the kinetic energy. The first one forces the robots to stay at a certain distance, $\delta$, from the object avoiding missing or colliding with it. We set $r_0 = \delta$ and $c_i c_j < 0$ in~(\ref{eq:cbpotential}). The second one indicates which velocity favors circumvent the object.
The strategy for moving around consists of computing a gradient of the points detected on the surface of the object and using it to sample velocities that favor the robot to move around the object.

\begin{figure}[!t]
		\centering
		\includegraphics[width=.80\columnwidth]{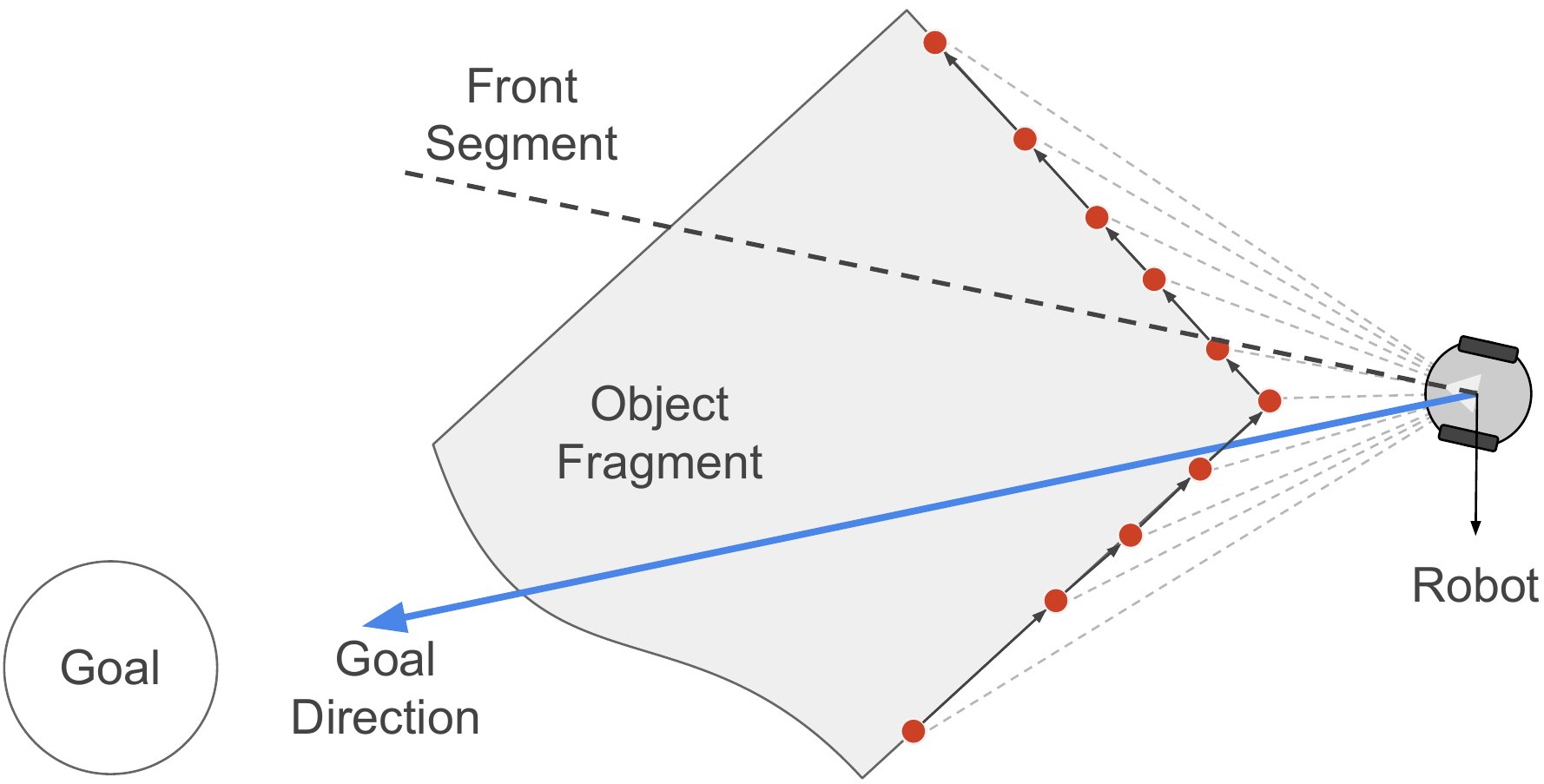}
	\caption{Illustrative scheme showing the perception of the robot regarding the object and its goal.} 
	\label{fig:occlusion_strategy}
\end{figure}

To compute the gradient, we arrange the points detected by the $i$-th robot on the $\mathcal{O}^{(t)}_i$ object on a clockwise or counterclockwise order. To decide in which order, we project a line segment (``front segment'' shown in Fig.~\ref{fig:occlusion_strategy}) and count the number of points to the left and to the right of that line segment. If there are more points to the left than right, we order clockwise; otherwise, we order them counterclockwise. This forces the robots to choose velocities that align them with the surface of the object. For now on, assume $\bar{\mathcal{O}}^{(t)}_i$ is the ordered set of points detected on the object. 

Defining the order of the elements directly implies on the direction of the gradient, which in turn indicates the direction that the robot should move around the object. Given $\bar{\mathcal{O}}^{(t)}_i$, the gradient is calculated as,
\begin{equation}
\nabla^{(t)}_i = \nabla\bar{\mathcal{O}}^{(t)}_i = \{ \mathbf{o}_{i+1} - \mathbf{o}_{i}: |\bar{\mathcal{O}}^{(t)}_i| \geq 2\},
\end{equation}
where $\nabla^{(t)}_i$ is a simplified notation and consists of a set with each element being a vector on the surface of the object.

To evaluate how much the velocity performed by the robot favors the move around behavior, we sum the differences between the gradient vectors and the robot velocity. That is, we compute the resulting vector,
\begin{equation}
\mathbf{Q}_i = \sum \limits_{\forall \mathbf{\upsilon_j} \in \nabla^{(t)}_i} \mathbf{\upsilon_j} - \bar{\mathbf{v}}_i, 
\label{eq:movearound}
\end{equation}
where $\bar{\mathbf{v}}_i$ is a sampled velocity for the $i$-th robot and $\mathbf{Q}_i$ is a vector that summarizes the mismatching among the sampled velocity and the gradient.


On the other hand, the {\em Pushing Behavior} is achieved simply by decreasing the $\delta$ distance applied to the Coulomb-Buckingham potential. This forces the robots to collide with the object producing the pushing behavior.
To achieve this behavior automatically, we use a similar mechanism presented previously to divide the points detected to the left or the right side regarding a line segment in the goal direction (shown as a blue arrow in Fig.~\ref{fig:occlusion_strategy}). If there are points on both sides, the object is occluding the goal from the robot; otherwise, the robot has a direct line of sight to the goal. The robot decides to start pushing the object if there are points on both sides and the proportion of points from one side to the other is higher than a threshold $\rho$. To start pushing the object, the robot must decrease the $\delta$ distance such that it collides with the object. If such a proportion is lower than $\rho$, then it increases the $\delta$ distance so that the robots can circulate the object.
\subsection{Obstacle avoidance}
To avoid collisions with obstacles, we extend the potential $U_{\{i\}}(\mathbf{v}_i)$ with another Coulomb-Buckingham potential. To compute such potential, we use distance $r$ between the points detected on the surface of obstacles and the \rev{robot\textquotesingle s pose}. By setting the charges in~(\ref{eq:cbpotential}) to ($c_i c_j > 0$), we note that the robots generate a repulsive behavior to obstacles avoiding collisions with them.

\subsection{Cohesive motion}
To cohesively navigate the swarm through the environment, we combine the Coulomb-Buckingham potential and the kinetic energy~(see Fig.~\ref{fig:diagram}). The first one forces the robots to aggregate, while the second one drives the robots to reach a consensus on the \rev{group\textquotesingle s velocity}.

To keep the cohesion of the swarm, we set the charges in~(\ref{eq:cbpotential}) to ($c_i c_j < 0$) and use the relative distances between the robot and its neighbors.

The velocity consensus is reached by computing the kinetic energy from the relative velocities of the $i$-th robot neighbors. That is,
\begin{equation}
\mathbf{V}_i = \sum \limits_{\forall \mathbf{v}_j \in \mathcal{N}_i} \mathbf{v}_j,
\end{equation}
where $\mathbf{V}_i$ is defined as the sum  of the relative velocities.

\subsection{Combination}
Formally, all the swarm behaviors described above and illustrated in Fig.~\ref{fig:diagram} are represented by the potential energy
\begin{equation}
\resizebox{0.45\textwidth}{!}{$%
    \begin{aligned}
            H_i(\mathbf{x}) = \left(
            \sum \limits_{\forall \mathbf{o_j} \in \mathcal{O}_i} \Phi(||\bar{\mathbf{q}} - \mathbf{o_j}||) + 
            \mathbf{E}(\mathbf{Q}_i)  + 
            \sum \limits_{\forall \mathbf{w_j} \in \mathcal{W}_i} \Phi(||\bar{\mathbf{q}} - \mathbf{w_j}||)\right) +
            \\ \left(
            \sum \limits_{\forall \mathbf{v_j} \in \mathcal{N}_i} \Phi(||\bar{\mathbf{q}} - \mathcal{K}(\mathbf{q}_j, {\mathbf{v}_j})||) +
            \mathbf{E}(\mathbf{V}_i) + \mathbf{E}(v_{max} - \bar{\mathbf{v}}_i)\right),
    \end{aligned}$%
    }
    \label{eq:paiwisepot}
\end{equation}
where $H_i(\mathbf{x})$ is the potential energy associated with the $i$-th robot concerning state $\mathbf{x}$ and a sampled velocity $\bar{\mathbf{v}}_i$. The first term in parenthesis defines, respectively, the cooperative transport and obstacle avoidance behaviors. The second one is related to the coordinated motion.

\begin{figure*}[!ht]
\centering
	\begin{subfigure}{.280\textwidth}
		\centering
		\includegraphics[width=\linewidth]{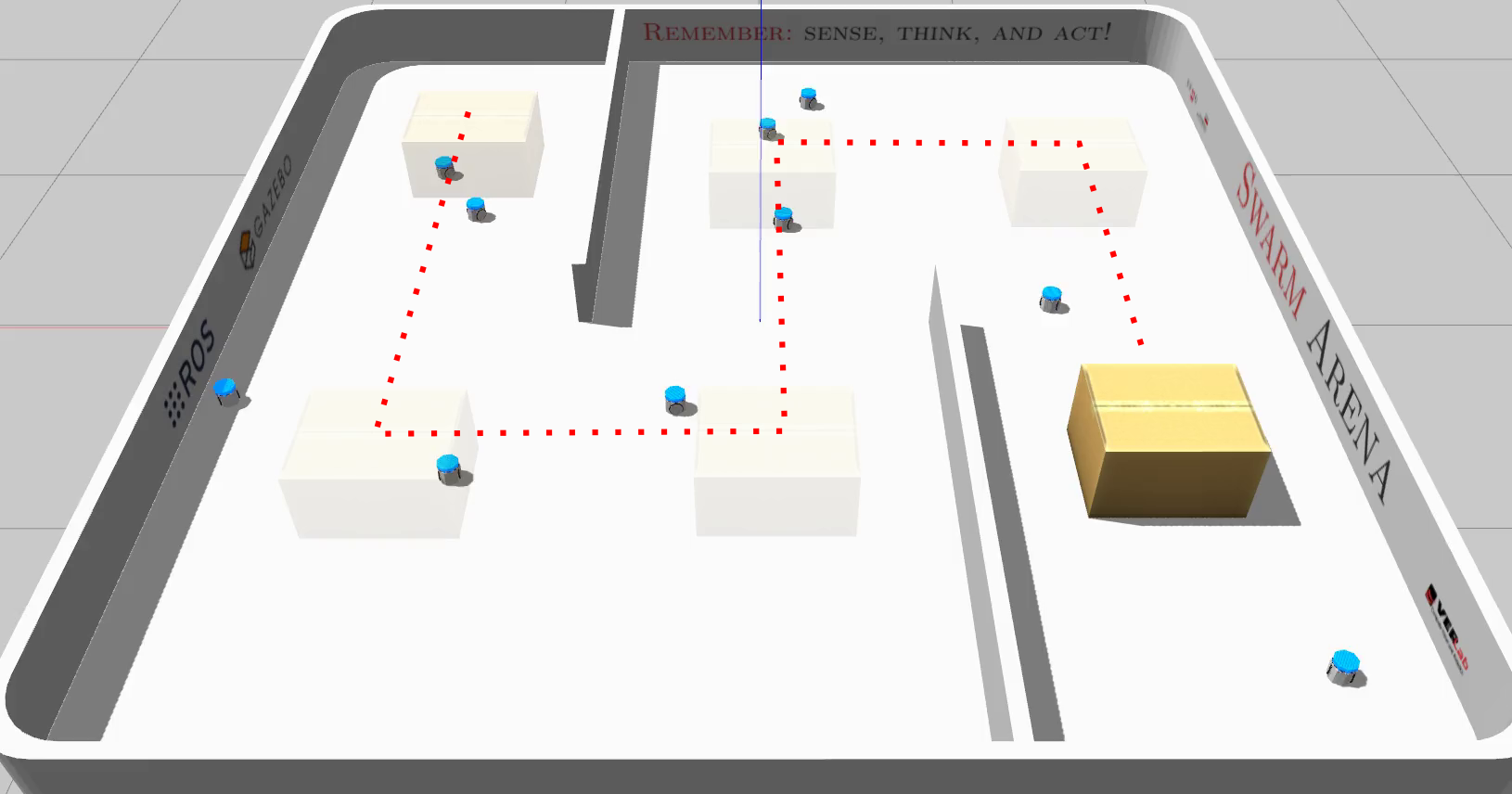}
		\caption{$t=0$ s}
	\end{subfigure}%
	\hfil
	\begin{subfigure}{.280\textwidth}
		\centering
		\includegraphics[width=\linewidth]{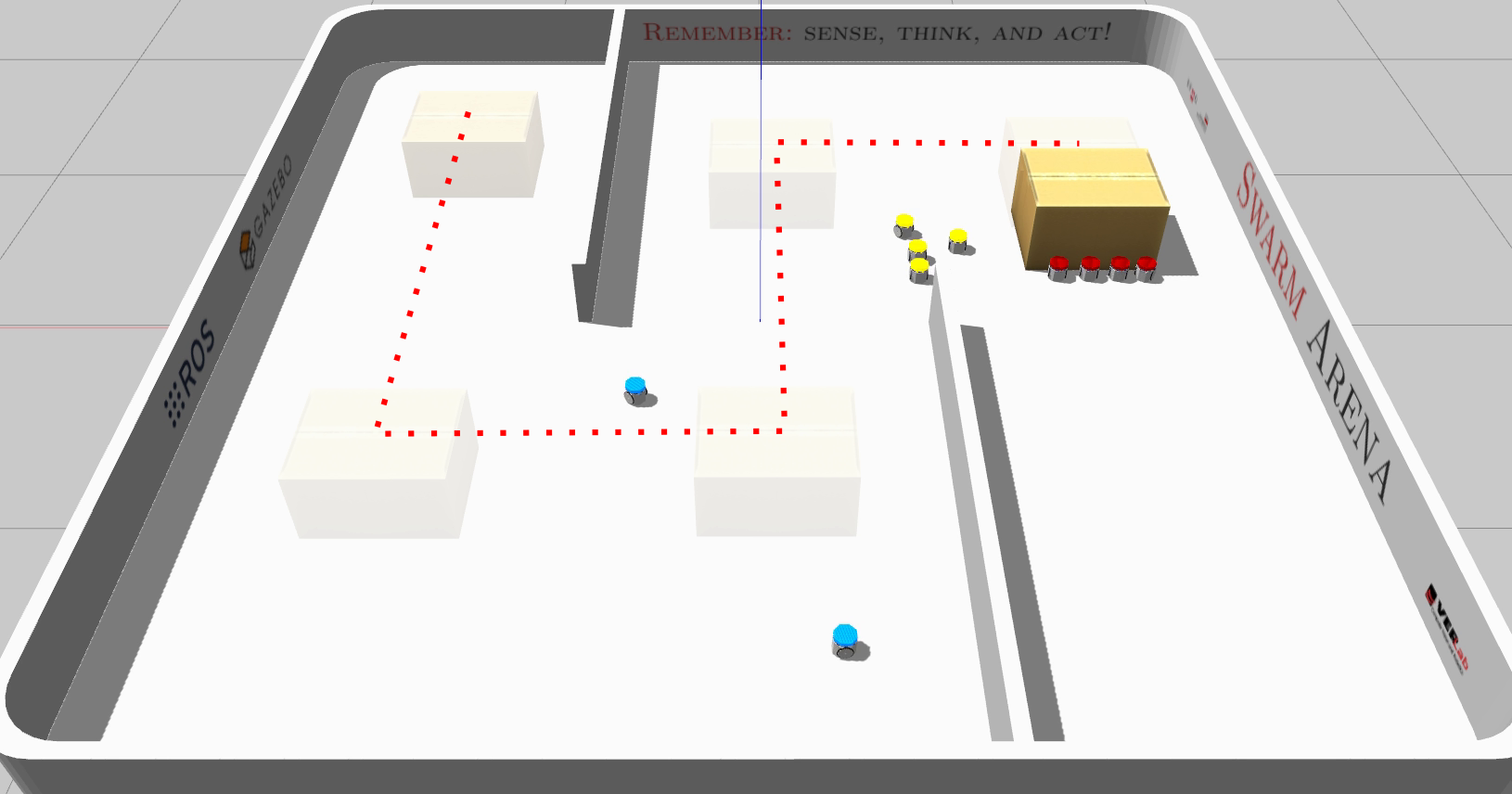}
		\caption{$t=338$ s}
	\end{subfigure}
	\hfil
	\begin{subfigure}{.280\textwidth}
		\centering
		\includegraphics[width=\linewidth]{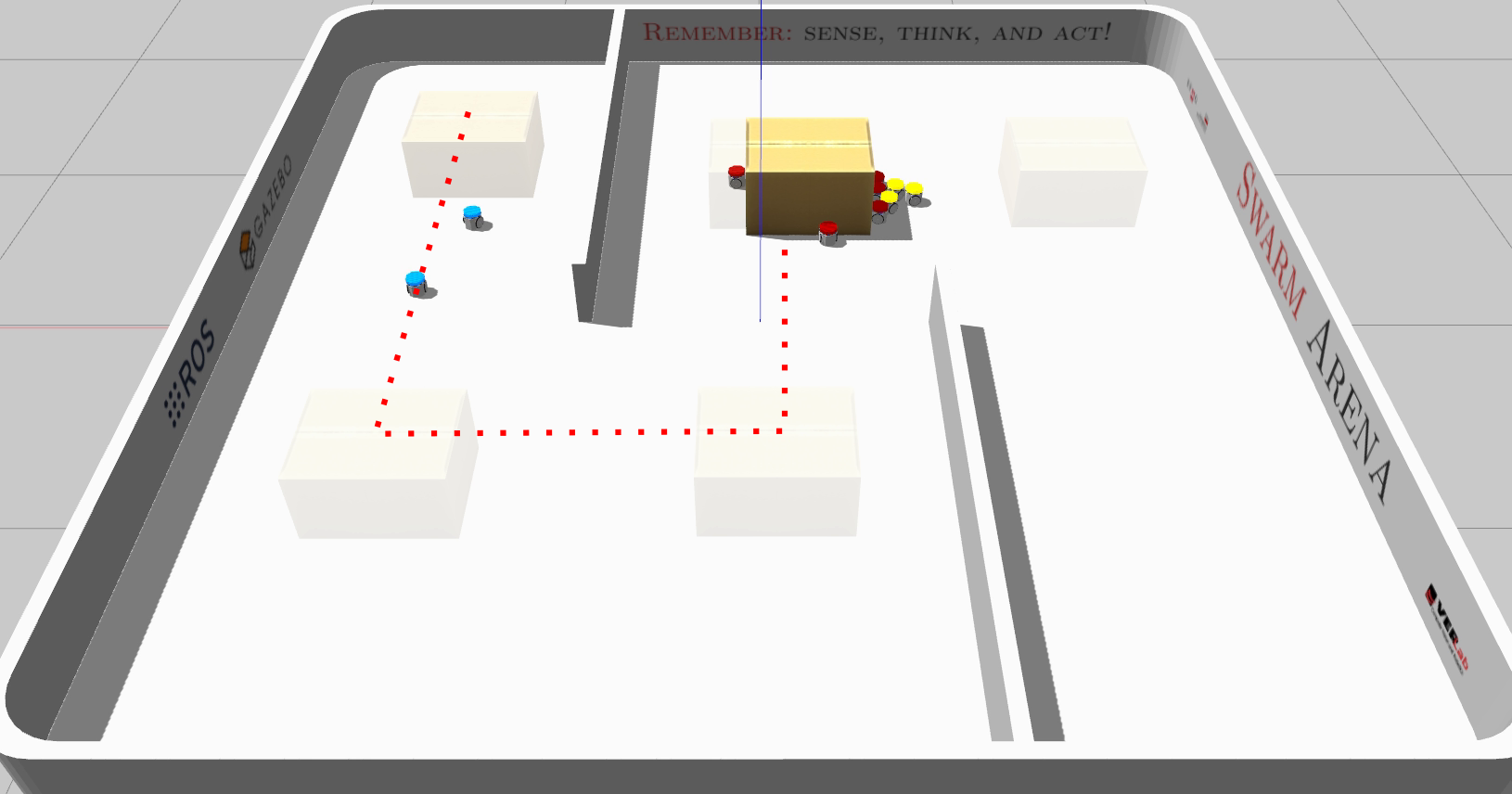}
		\caption{$t=485$ s}
	\end{subfigure}
	\medskip
	\begin{subfigure}{.280\textwidth}
		\centering
		\includegraphics[width=\linewidth]{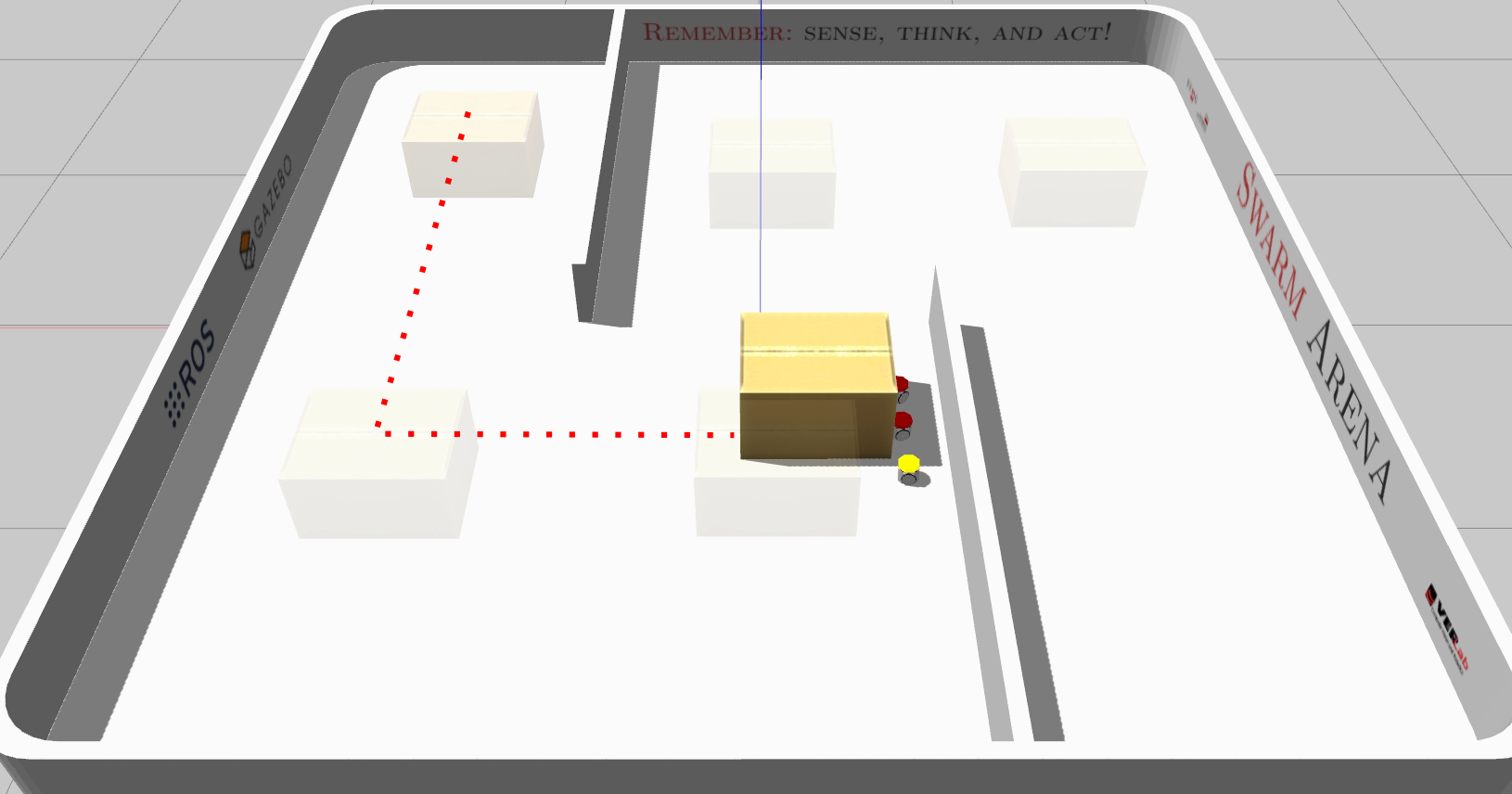}
		\caption{$t=642$ s}
	\end{subfigure}
	\hfil
	\begin{subfigure}{.280\textwidth}
		\centering
		\includegraphics[width=\linewidth]{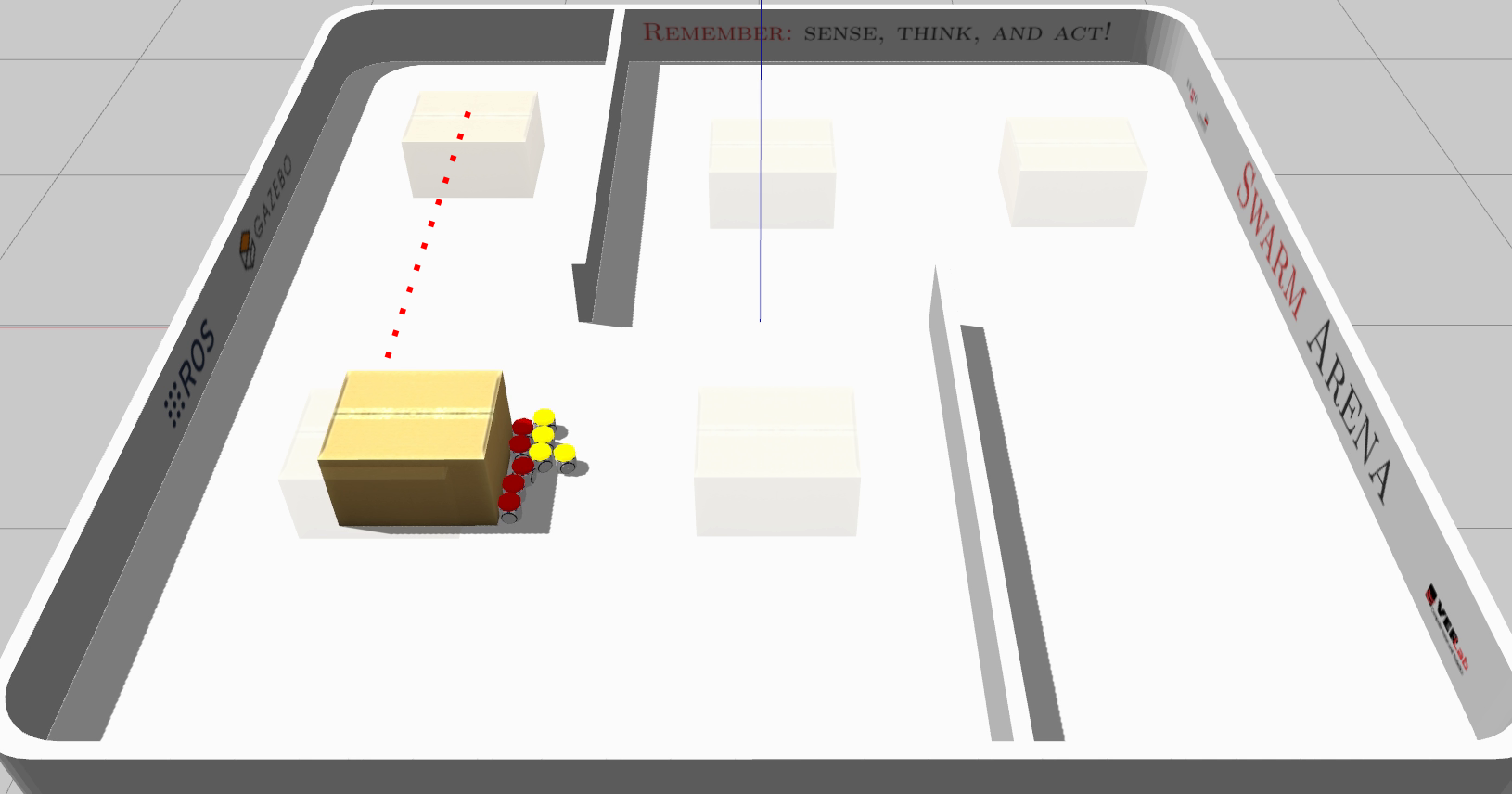}
		\caption{$t=805$ s}
	\end{subfigure}
	\hfil
	\begin{subfigure}{.280\textwidth}
		\centering
		\includegraphics[width=\linewidth]{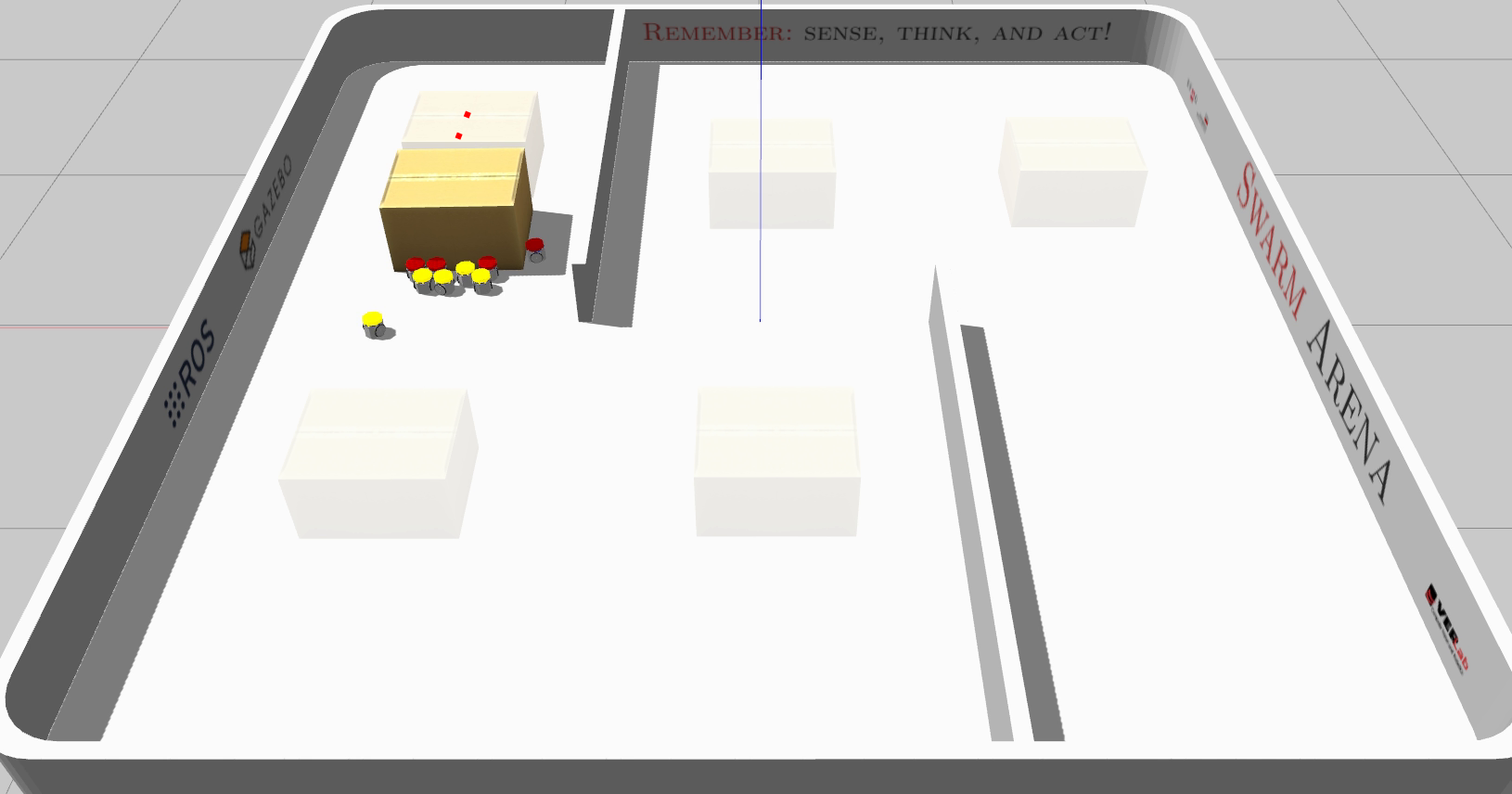}
		\caption{$t=919$ s}
	\end{subfigure}
	\caption{Snapshots of an experiment showing $10$ simulated robots transporting an object toward a sequence of goal locations in a complex environment.}
	\label{fig:navigation}
\end{figure*}

\begin{figure*}[!ht]
\centering
	\begin{subfigure}{.220\textwidth}
		\centering
		\includegraphics[width=\linewidth]{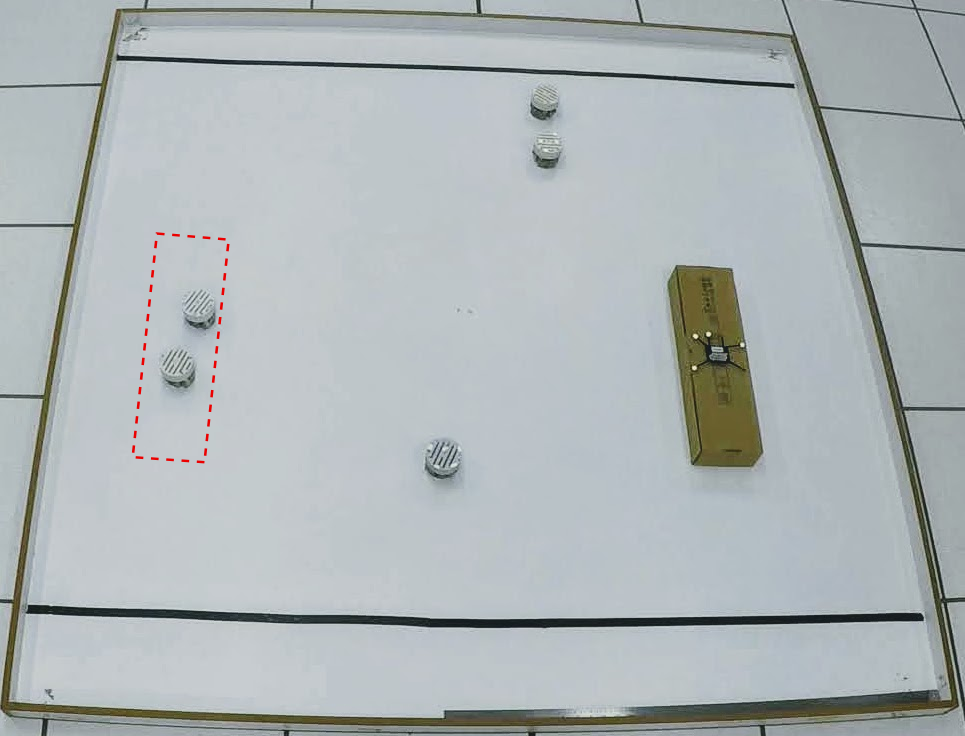}
		\caption{$t=0$ s.}
	\end{subfigure}%
	\hfil
	\begin{subfigure}{.220\textwidth}
		\centering
		\includegraphics[width=\linewidth]{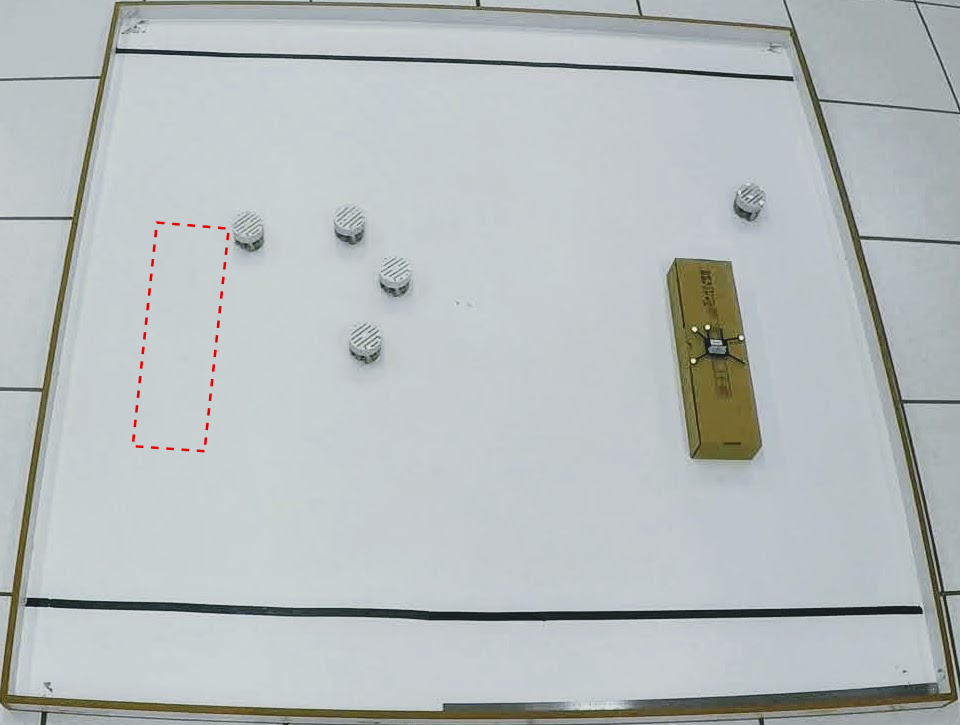}
		\caption{$t=65$ s.}
	\end{subfigure}
	\hfil
	\begin{subfigure}{.220\textwidth}
		\centering
		\includegraphics[width=\linewidth]{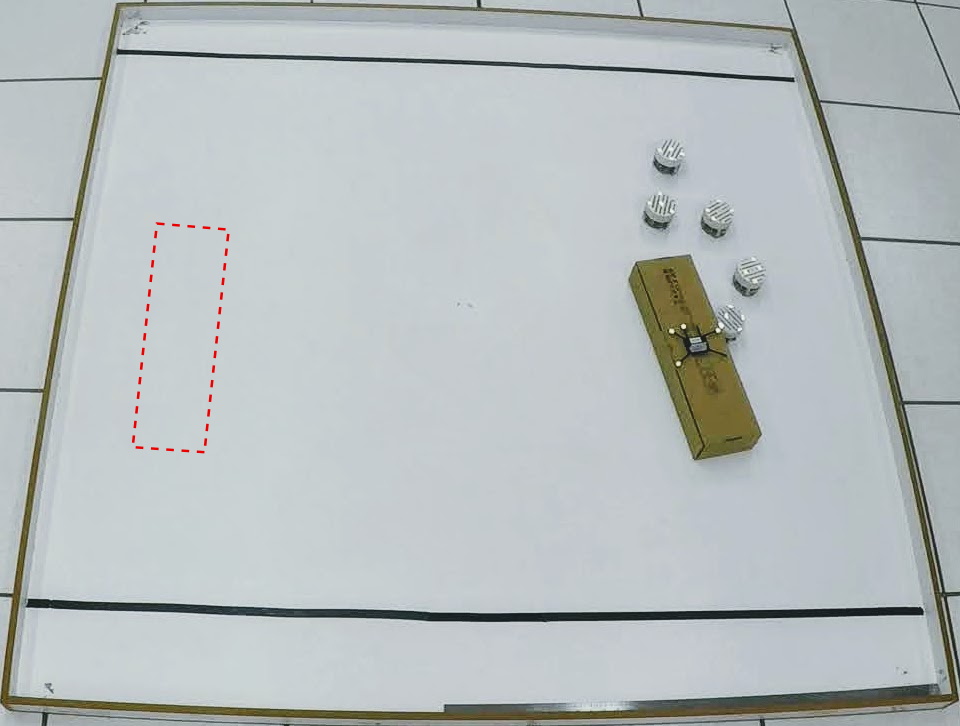}
		\caption{$t=146$ s.}
	\end{subfigure}
	\hfil
	\begin{subfigure}{.220\textwidth}
		\centering
		\includegraphics[width=\linewidth]{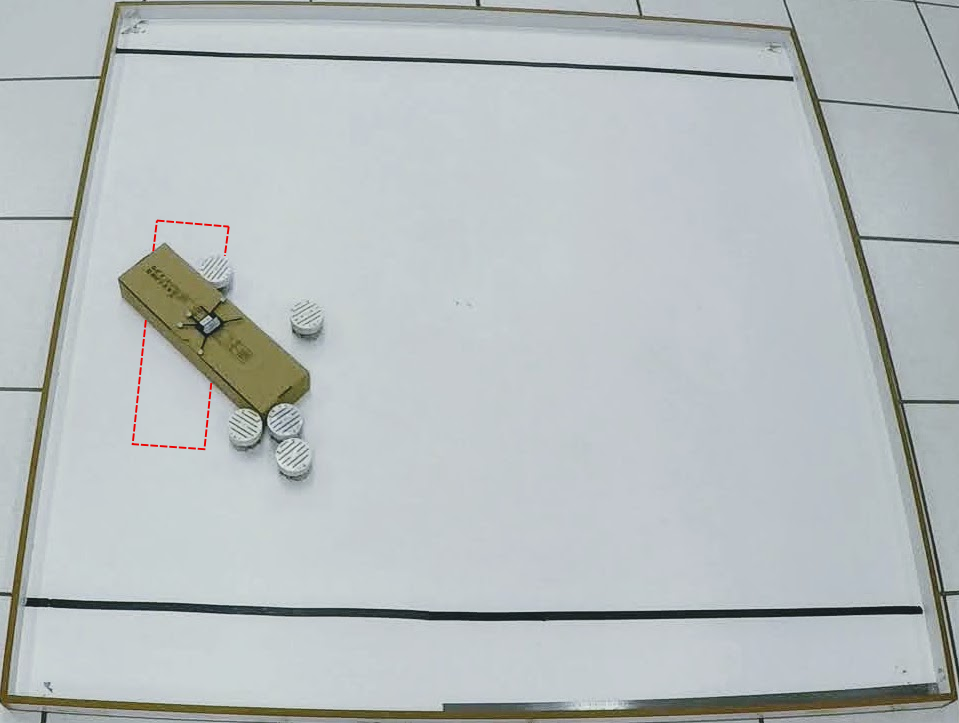}
		\caption{$t=307$ s.}
	\end{subfigure}
	\hfil
	\caption{Snapshots of an experiment showing $5$ e-puck robots transporting an object toward its goal. The red rectangle indicates the goal location.}
	\label{fig:real}
\end{figure*}

\section{Experiments and Results}
\label{sec:experiments}
We conduct several simulated experiments to evaluate our approach's performance regarding scalability, adaptability, and robustness. For this, we implemented our methodology using the ROS (Robot Operating System)\footnote{\url{https://www.ros.org}} middleware and set up a simulated environment in Gazebo\footnote{\url{http://gazebosim.org}}. The simulated environment consists of a $4$ by $4$~meters arena, an object, and a group of differential drive robots. The simulated robots are a model of a physical robot that we are developing in our laboratory for experiments with swarm robotics~\cite{rezeck2017hero}. It is equipped with a range sensor consisting of a set of \rev{infrared} beams with $\lambda = 0.5$~meters. We consider that this sensor can distinguish between objects and obstacles. As mentioned, we also consider that robots can estimate neighbors' positions and velocities. The maximum speed reached by these robots is $0.12$~m/s. The simulator's physics engine allows a single robot to start the movement of a $200$~grams object if it collides with it at speeds above $0.10$~m/s. It also allows it to keep pushing the object if it maintains a speed greater than $0.01$~m/s. 
For each experiment, we performed $30$ runs and, for each run, the robots are randomly placed in the environment. The results are presented as an average value with a $95\%$ confidence interval. \rev{A video of the experiments is available on Youtube\footnote{\url{https://youtu.be/hrkJKL3W3pQ}} and the source code at Github\footnote{\url{https://github.com/rezeck/grf_transport}}}.

\subsection{Scalability}
\begin{figure}[!ht]
	\centering
	\includegraphics[width=0.95\linewidth]{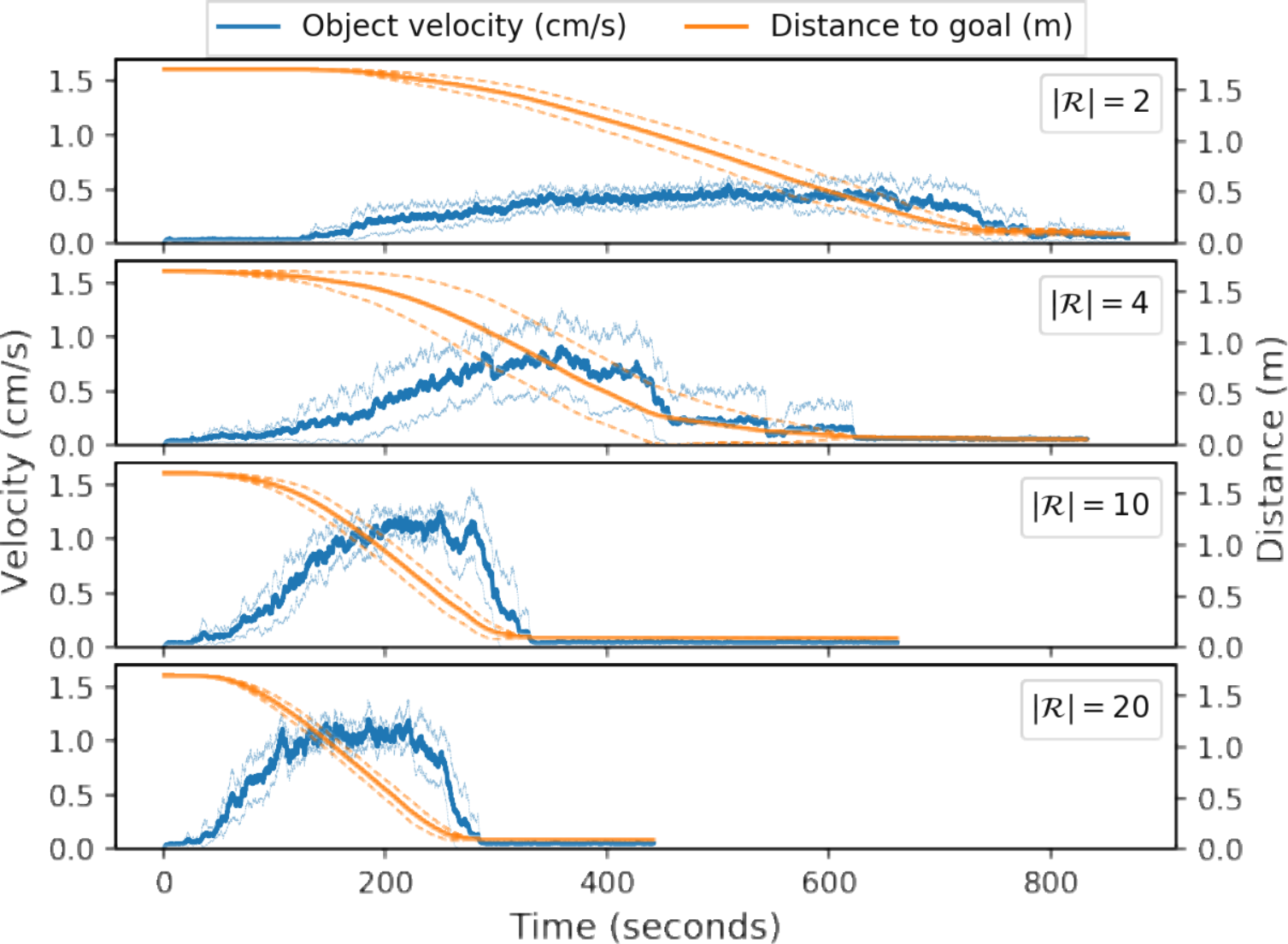}
	\caption{Experiments on the scalability of the system when increasing the number of robots. 
	}
	\label{fig:scalability}
\end{figure}

To evaluate the system scalability, we performed experiments increasing the number of robots: $2$, $4$, $10$, $20$. A rectangular object of $0.5$ by $0.4$~meters and mass equal to $200$~grams was placed in the center of the arena and the goal is located $1.7$~meters ahead. As metrics, we measure the velocity and current distance between the object and its goal location. We assume the robots stop pushing the object if it is less that $0.1$~meters of its goal. Fig.~\ref{fig:scalability} shows the results from this experiment and Table~\ref{tab:scalability} summarizes the average time spent by the robots to transport the object to its goal.

As expected, increasing the number of robots reduces the time taken to find and transport the object to the goal location. With more robots searching, the time taken to find the object is smaller. Moreover, the object reaches higher speeds when pushed by more robots, also reducing the transport time.

However, we also observe that the gain is smaller for a large number of robots. For example, the gain of increasing from $2$ to $4$ robots is larger than from $10$ to $20$. When there are too many robots trying to push the object, the most distant ones cannot detect it due to sensing restrictions and will not contribute to the pushing. Thus, while our approach is scalable, the performance will depend on the size of the object and robots' sensing capabilities, and it may saturate for a large number of robots.

\begin{table}[!ht]
\caption{Average transportation time \rev{for 30 trials} when the number of robots increases.}
\label{tab:scalability}
\centering
\begin{tabular}{cc}
\toprule
\textbf{Number of robots} & \textbf{Transport time (seconds)} \\ \midrule
$2$ & $829 \pm74$ \\
$4$ & $593 \pm96$ \\
$10$ & $324 \pm18$ \\
$20$ & $278 \pm14$ \\ \bottomrule
\end{tabular}
\end{table}

\subsection{Adaptability}
We propose two experiments to determine whether our approach allows the swarm to adapt to robot failures and changes in the environment. We use $10$~robots in these experiments, and we move a rectangular object of $0.5$ by $0.4$~meters with a mass equal to $200$~grams toward its goal,  located $2.6$~meters apart. Unlike the previous setup, we place the object such that it requires the robots to push it around corners. 

\begin{figure}[t]
	\centering
	\includegraphics[width=0.95\linewidth]{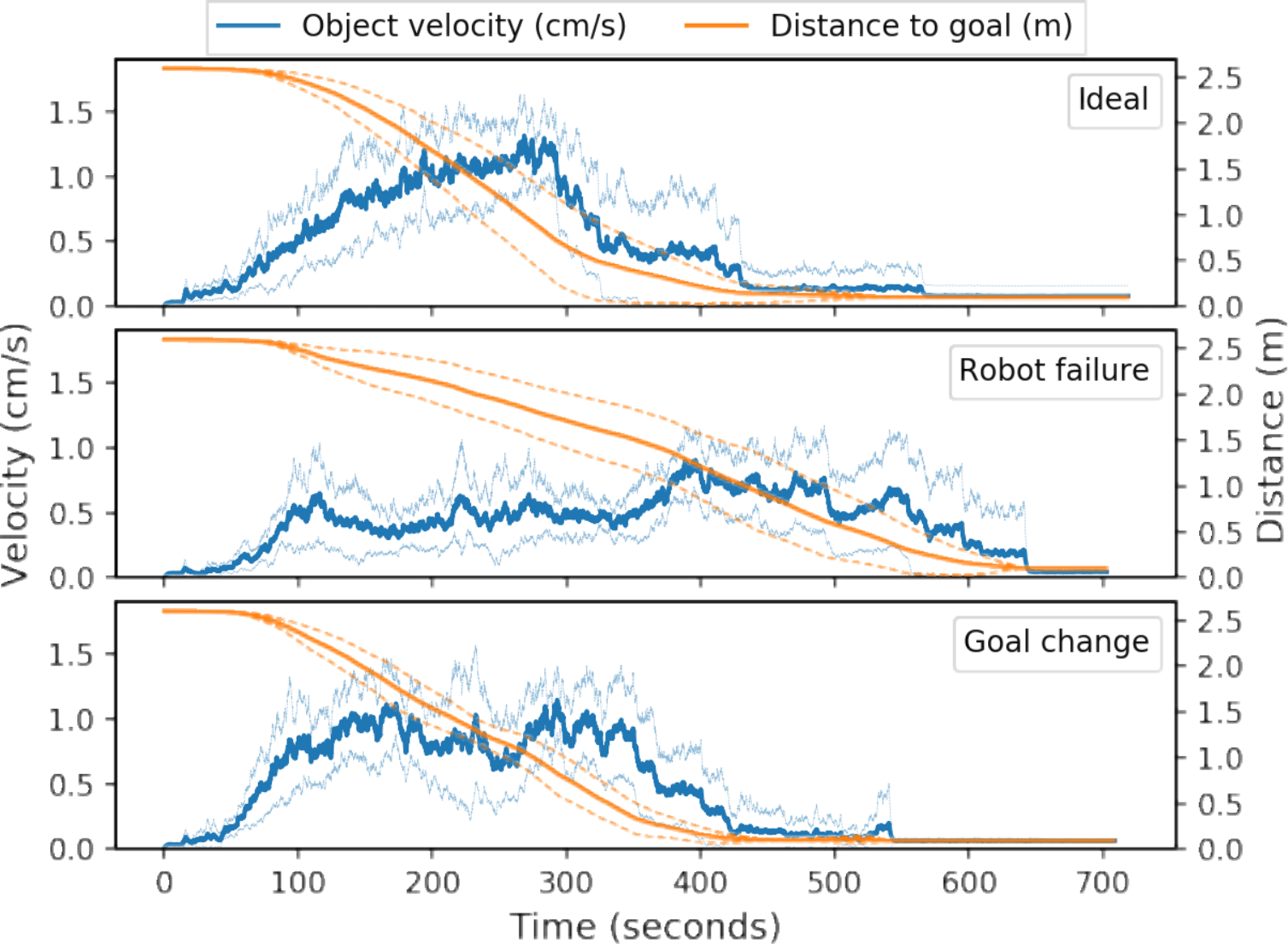}
	\caption{Experiments on the adaptability of the swarm when robot failure occurs and when the goal location changes. 
	}
	\label{fig:adaptability}
\end{figure}

In the first experiment, we assume that $4$ robots stop working due to mechanical failures when they are pushing the object. We intend to evaluate the method's resilience when the number of robots performing the task abruptly decreases. Also, this experiment allows us to assess the robots' resilience to execute the task. Fig.~\ref{fig:adaptability} shows the average velocity and distance of the object from its goal. Table~\ref{tab:adaptability} summarizes the average time spent by the robots to transport the object.

In Fig.~\ref{fig:adaptability}, we may observe that robot failures impact the duration of the task. Besides the impact of having fewer robots pushing the object, we noticed that in many runs, the ``dead" robots obstructed the others and were also pushed together with the object, decreasing the object velocity. 
Despite that, we noticed that the swarm managed to successfully transport the object in all attempts.

In the second experiment, we evaluate how the swarm behaves when the goal location is changed. To assess this, we changed the goal when the object is less than $1.3$~meters from its goal. We set the new goal location to $1.3$~meters in the other direction so that the robots need to move around the object to continue the transport. The total distance with the change of the goal is $2.6$~meters. In spite of that, Figure~\ref{fig:adaptability} shows that the robots are able to complete the task, and the time taken does not increase significantly, demonstrating the adaptability of the method. 

\begin{table}[!t]
\caption{Average transportation time \rev{for 30 trials} when the robots need to adapt to changes in the environment.}
\label{tab:adaptability}
\centering
\begin{tabular}{cc}
\toprule
\textbf{Scenario} & \textbf{Transport time (seconds)} \\ \midrule
Ideal & $438 \pm 13$ \\
Robot failure & $596 \pm 21$ \\
Goal change & $449 \pm 11$ \\ \bottomrule
\end{tabular}
\end{table}

\subsection{Robustness}
\begin{figure}[t]
	\centering
	\includegraphics[width=0.95\linewidth]{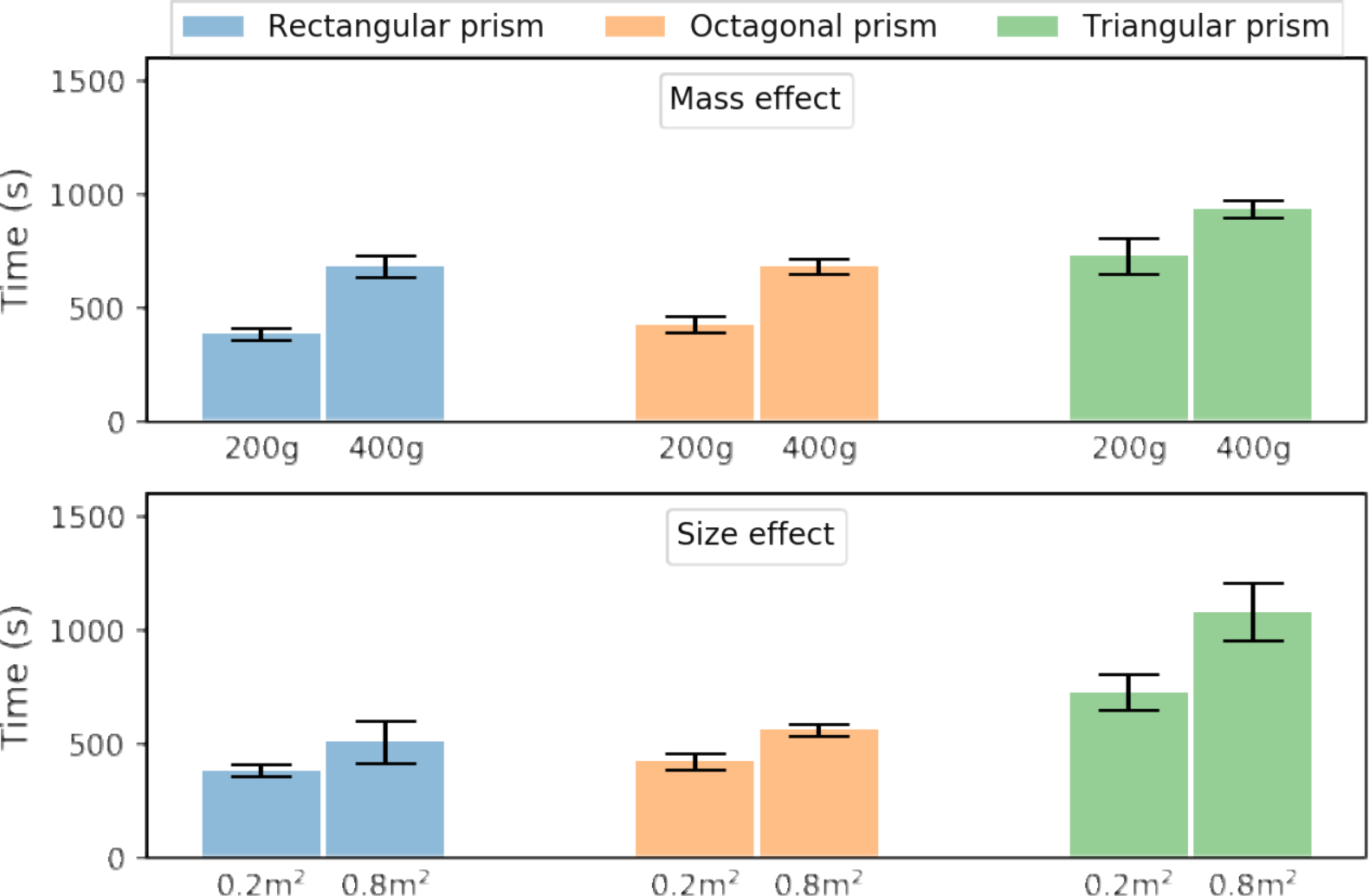}
	\caption{Experiments on the robustness of the swarm regarding different objects with different sizes and masses.}
	\label{fig:robustness}
\end{figure}

We conducted experiments with three objects of different shapes, sizes, and masses to assess the robustness of the method. To evaluate how robots behave when they detect objects with differently shaped corners, we used three objects: a rectangular prism (right angle), an octagonal prism (obtuse angle), and a triangular prism (acute angle). For each object, we experiment the effects of doubling their sizes and masses. In all experiments, we placed the object at a distance of $2.6$~meters from its goal. Fig.~\ref{fig:robustness} shows the time required to transport the objects to the goal.

In this experiment, the robots were less efficient in pushing the triangular prism than the other two objects. When the robots start pushing the triangle at its acute corners, few robots are grouped to push, increasing the task time. As expected, increasing the mass of the object also increases the transport time. \rev{Although larger objects enable more robots to push it, they also cause an overall increase in transport time. Increasing the effective contact surface between the object and the ground increases the maximum intensity of the friction force, making the object more difficult for the robots to push.} When we increase the size of the triangular prism, we observe that its impact over the transport time is relatively higher than increasing either the rectangular or the octagonal prism. We believe that objects with sharp corners decrease our method's performance since we assume our sensor to be radial, making detection difficult when the robot is pushing on a corner. Despite the disparity in the transport time for different objects, the robots successfully transported all objects to their goals.

\subsection{Object transportation in complex environments}
In more complex environments, pushing an object toward its goal location may be challenging due to obstructions caused by obstacles. Navigation strategies can help robots to move an object to its final location, passing through a series of waypoints.

Since our method allows robots to adapt to the change in goal location, we conducted experiments where the robots transport an object through a sequence of goals. We assume the robots are initially distributed around the environment, do not know the object's location, but have the sequence of waypoints beforehand. Fig.~\ref{fig:navigation} displays a sequence of snapshots from one experiment and a video\footnote{\url{https://youtu.be/a0eYZid3jQs}\label{foot:navvideo}} is also  available.

\subsection{Real robots}
We also conducted proof-of-concept experiments to assess the applicability of our approach in a real environment. The environment consists of a bounded square area of $2$ by $2$~meters, an object of $0.14$ by $0.48$~meters \rev{that weighs $550$~grams}, and five e-puck robots~\cite{mondada2009puck}. The robots receive velocity commands from a remote server executing ROS, where each one of them is implemented as an independent node. Given that our robots are not equipped with sensors that enable them to process our method, we emulate such sensors using the information provided by an Optitrack motion capture system\footnote{\url{http://optitrack.com}}. We set the sensing distance to $\lambda = 0.3$ and set the goal location $1.1$~meters ahead of the object.

We performed $10$ runs with a duration of $10$ minutes each. As we expected, the robots were able to transport the object in all attempts, showing that the method is effective even with the uncertainties found in real scenarios. Overall, the robots took $448 \pm 8$ seconds to transport the object. Fig.~\ref{fig:real} presents a sequence of snapshots from one experiment and a video\footnote{\url{https://youtu.be/1I9-hTQO8CU}\label{foot:real}} is also available. 

%
\section{Conclusion and Future Work}
\label{sec:conclusion}
This paper presented a novel approach based on Gibbs Random Fields that allows a swarm of robots to navigate through an environment and cooperatively transport an object toward its goal location. The approach is fully decentralized and based on local information, being suitable for robotic swarms.
To assess the scalability, adaptability, and robustness of our method, we conducted several experiments in different scenarios. Results showed that our method is scalable and supports the transportation of objects of different shapes, sizes, and masses. Moreover, it exhibited resilience to changes in goal location and robot failures. Proof-of-concept experiments using real robots showed the feasibility of our approach in a real-world environment. By setting a sequence of goals, we further demonstrated our method's application in more complex environments, where the robots push an object through an environment with obstacles.
In future work, we intend to explore heterogeneity to allow part of the group to take responsibility for finding the goal location and indicating it to other robots in charge of transporting the object. Also, \rev{we plan to experimentally compare our work with others in the literature} and investigate ways to allow our method to adjust the object's orientation regarding its goal.







\bibliographystyle{unsrt}
\bibliography{root}

\end{document}